# Deep Radar Detector

Daniel Brodeski[1,2], Igal Bilik[1] and Raja Giryes[2]
[1] General Motors - Advanced Technical Center Israel
[2] School of Electrical Engineering, Tel-Aviv University
Emails: {daniel.brodeski, igal.bilik}@gm.com, raja@tauex.tau.ac.il

*While camera and LiDAR processing have been revolutionized since the introduction of deep learning, radar processing still relies on classical tools. In this paper, we introduce a deep learning approach for radar processing, working directly with the radar complex data. To overcome the lack of radar labeled data, we rely in training only on the radar calibration data and introduce new radar augmentation techniques. We evaluate our method on the radar 4D detection task and demonstrate superior performance compared to the classical approaches while keeping real-time performance. Applying deep learning on radar data has several advantages such as eliminating the need for an expensive radar calibration process each time and enabling classification of the detected objects with almost zero-overhead.*

## I. Introduction

Autonomous driving is one of the major industrial trends that is expected to affect our lives in the near future. One of its main challenges is perception, i.e., understanding the surrounding. To address this challenge, autonomous vehicles use a set of sensors (e.g. cameras, LiDARs, and radars) to "sense" their surroundings and a set of algorithms to build the 3D world representation. Recently, *deep learning* (*DL*) became the key component of these perception techniques [2, 4, 5, 19]. It emerges as the main enabler to the great progress made recently in fields such as image recognition [1,2], object detection [3,4], instance segmentation [4] and 3D object detection [5]. While deep learning is applied mainly on the camera and LiDAR sensors data, radar processing still relies mostly on "classical tools". In this work, we wish to take a step towards making *DL* more applicable to radar processing.

The recent demand for autonomous driving drove the automotive industry towards a new generation of high-resolution (azimuth and elevation) automotive "imaging" radars [6, 7]. The main goal of these imaging radars is to create a relatively dense point cloud (less dense than lidars) of the vehicle surrounding at a lower cost and with a superior weather immunity compared to optical LiDARs. One may expect that the radar data would generate a LiDAR-like point cloud and that high-level algorithms, such as object detection and classification, would be adopted easily from the LiDAR pipeline. In reality, radar-generated point clouds differ significantly from the LiDAR point clouds in two aspects:

- Viewpoint/pose variation – a point cloud of an object differs for even very similar object poses and close viewpoint angles.

- Temporal-variation – even with no pose variation, a point cloud of the same object vary over time.

This research is partially supported by ERC-StG grant no. 757497 (SPADE).

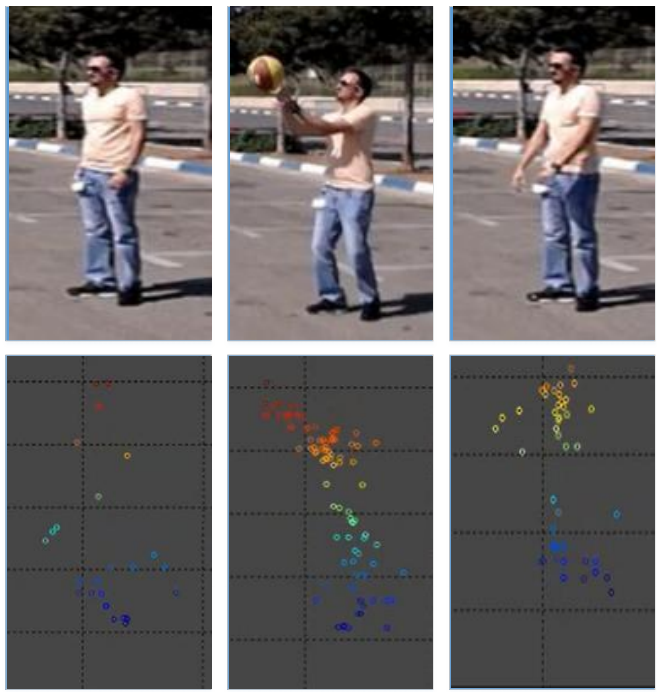

Fig. 1. The same object (person) in 3 similar cosequitive poses/view points generate significantly different radar point clouds (side-view).

Representative example in Fig. 1 demonstrates the variation of the object point cloud over a few consecutive frames. These variations occur both due to the radar sensors statistical nature and due to the use of the "classical" radar processing, which uses "handcrafted features" (e.g. detector thresholds) to detect and localize radar detection.

Motivated by the recent success of DL in computer vision, it has been used for static object classification using radar data [10]. In [11], DL is used to solve the problem of radar-based classification of aided and unaided human activities. A radar-based fall motion detection using DL has been proposed in [12]. All these works use occupancy grid or spectrograms as the image representation of the radar data for object classification. DL radar processing using non-image-like data is proposed in [13] for cognitive radar antenna selection, where the input to the network is a set of features from the estimated covariance matrix. However, also here the network task remains a classification task. Schumann, Ole et al [22] applied DL for semantic segmentation of radar point clouds but does it on a preprocessed radar point cloud. To the best of our knowledge, a DL method for detection and localization of multiple objects from various classes in a single complex frame is currently missing.

This work introduces the deep radar detection (DRD) approach for radar data processing using a convolutional neural network (CNN). A significant challenge of applying DL to radar data is the lack of labeled data. This work proposes an innovative way to extract labeled radar data from the radar calibration process and use it for training by introducing a novel data augmentation method for it.

The proposed method uses the complex (real and imaginary) radar data directly, and therefore, is able to detect and localize in the 4D space of range, Doppler, azimuth and elevation, multiple detections from multiple classes. The proposed DRD approach aims at replacing the entire "classical" radar signal processing blocks of detection and beam-forming. We show that it outperforms the classical radar processing performance while maintaining real-time abilities. Moreover, the proposed approach eliminates the need for the expensive and time-consuming radar calibration process (except for the initial training process). Finally, our method allows classifying the objects detected by the radar with no extra computational overhead. This provides new abilities that are not offered by the classical blocks of detection and beamforming.

The rest of the paper is organized as follows. Conventional radar signal processing is summarized in Section II. The proposed DRD approach is described in Section III. Section IV evaluates the performance of the proposed DRD technique, and Section V concludes the paper.

## II. CONVENTIONAL RADAR SIGNAL PROCESSING

Fig. 2 shows the conventional radar signal processing flow. The sampled radar echoes are first transferred to range-Doppler (RD) domain via the 2D fast Fourier transform (FFT). Next, the radar signals in the RD map, whose energy exceeds the detection threshold are declared as detections. In the following beamforming processing block, the direction in azimuth and elevation to these detections is estimated. Finally, detections are clustered, tracked and classified. Notice that the quality of the radar point cloud is mainly determined by the detector and beamforming.

Cell averaging-constant false alarm rate (CA-CFAR) detector is effectively used in the conventional radar processing [9]. For each RD-map cell under test (CUT), the energy in the surrounding cells $Z_k, k = 1, ..., M$, excluding the guard cells, is averaged to obtain the detection threshold:

$$y = \frac{1}{M}\sum_{k=1}^{M} Z_k,  \quad (1)$$

The detection in the CUT is then declared if its energy exceeds this detection threshold:

$$\begin{cases} CUT > \alpha y, \text{ detection} \\ CUT \leq \alpha y, \text{ not detection} \end{cases} \quad (2)$$

The direction of arrival (DOA) of all detections in the RD-map are obtained via beamforming (BF) [8]. Multiple beamforming methods, such as Bartlett and minimum variance distortion less response (MVDR) can be used in automotive multiple-input-multiple-output (MIMO) radars.

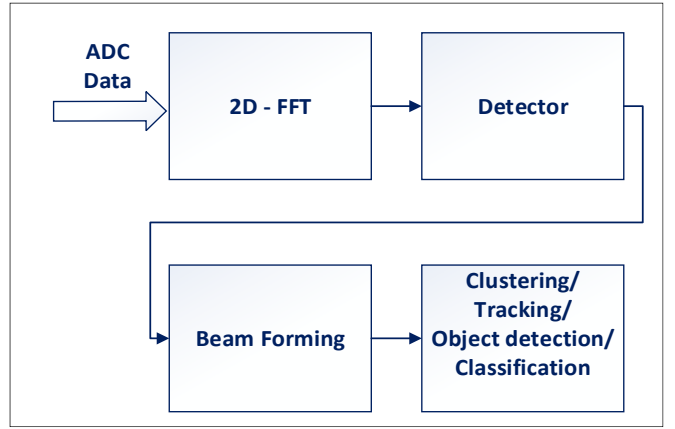

Fig. 2. Conventional radar signal processing.

Beamforming requires sensor array calibration, where array responses to targets at various known positions are collected to construct a sensor array calibration matrix.

Notice that multiple user-defined parameters, such as threshold, margin, sizes and shapes of the reference and guard windows determine the performance of the conventional radar signal processing. Automotive radar is required to operate in a variety of significantly different scenes, and therefore, selection of a single optimal set of these parameters is extremely challenging and frequently an impossible task. The main goal of this work is to develop a data-driven DL approach for radar signal processing and thus to eliminate the need for an accurate parameters selection.

## III. DEEP RADAR DETECTION

### A. Data

The availability of a labeled dataset is considered as a mandatory requirement for employing DL to a certain application. While there are multiple, publicly available datasets for RGB images and LiDAR [15, 16, 17], there are very few ones for radar. Moreover, even the available ones, e.g., the recently published nuScenes dataset [18], do not have the sampled radar ADC data required for deep radar detection.

One possibility to overcome the lack of labeled data is the syntactic data generation method that relies on some sensor model. Unfortunately, radar modeling is extremely challenging and computationally demanding due to the multiple effects that occur in acquisition time such as multipath reflections, interference, reflective surfaces, discrete cells, and attenuation. Another possibility is to use the DL approach for synthetic automotive radar scenes generation as shown in [14]. Yet, as it is a simulation based method, it also relies on a world model that may include inaccuracies.

This work proposes to use the radar calibration data, which contains the radar sensor array responses to a known target located at a variety of angles. Typically, the radar is calibrated in the anechoic chamber with a known point-target (corner reflector). The radar is mounted to an accurate rotator to collect array responses at a variety of angles. Fig. 3 shows a typical anechoic chamber being used for radar calibration.

Fig. 4 (left) shows the raw radar frame, collected during the calibration process along with the ground-truth label of the detection: the range, Doppler, and azimuth and elevation angles. Dimensions of the raw radar frame are $N_s \times N_c \times N_{Ant}$, where $N_s$ is the number of samples, $N_c$ is the number of chirps and $N_{Ant}$ is the number of receiver antennas or the number of virtual elements in the MIMO array (in a TDM based MIMO). The output of the 2D-FFT, applied to the raw radar frame has the size of $N_R \times N_D \times N_{Ch}$, where $N_R$ is the number of range-bins, $N_D$ is the number of Doppler-bins and $N_{Ch}$ is the number of channels. It is used as an input to the DRD network (Fig. 4 right). Notice that the radar data is complex, and therefore, each channel is represented via the real and imaginary parts, resulting in $N_{Ch} = 2N_{Ant}$.

The proposed approach for the radar calibration data suffers from the lack of diversity (radar echoes from only two static targets at fixed ranges and Dopplers). Therefore, we use data augmentation with the raw radar data to address this limitation. The phase of the radar echo contains information on the target range and Doppler. It may be represented using the following linear frequency modulation (LFM) signal (3):

$$x_{LFM}(t) = e^{j\pi\alpha t^2}, \quad (3)$$

where $\alpha = B/T_{Chirp}$ is the frequency modulation of the chirp. Here, $B$ is the signal bandwidth (BW) and $T_{Chirp}$ is the chirp duration. The LFM signal can be shifted in range and doppler using (4):

$$x_{augment}(t) = x_{LFM}(t) \cdot e^{j\pi a_m(t)}, \quad (4)$$

where $a_m(t)$ is the augmentation phase shift we want to generate. Therefore, radar targets at any range and Doppler can be augmented simply by shifting the phase of the raw radar frame elements. We do that easily by multiplying window coefficients of the 2D-FFTs with a complex exponent (before applying the FFT on the data, we first pass it through a window (e.g. hamming) to reduce side-lobes).

### B. Network Architecture

Fig. 5 shows the proposed network flow whose main goal is to detect and localize in the 4D space (range, Doppler, azimuth and elevation) multiple objects in the raw radar frame provided at the input. The network architecture can be found in Fig. 6.

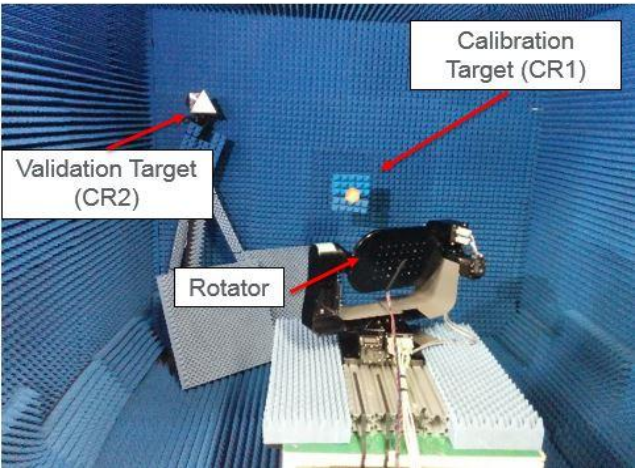

Fig. 3. Radar Calibration Process

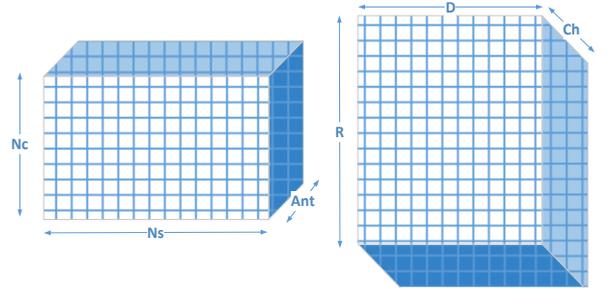

Fig. 4. Raw Radar Frame (left); Network Input Radar Frame (right)

Our solution relies on a two-step detector as in the faster-RCNN (Region CNN) model [3,4], whose detection is performed using the following two steps:

- Region Proposal Network (RPN) propose regions where it is likely to find objects. Each is provided with its possible coarse location (using anchors).

- Classifier - classifies the proposed objects and finetune their locations (via regression).

Our model proposes the following two detection steps:

- RD-Net: detects, classifies and localizes all detections in the range-Doppler domain.

- Ang-Net: finds the azimuth and the elevation angles of each detection found by the RD-Net.

In this work, the detection task in the Range-Doppler map is formulated as a segmentation, in which each cell ("pixel") in the RD-map is labeled by the correct class. We found this segmentation approach to be more effective than [3, 4] for our problem. The RD-Net, whose internal architecture adopts the 2D-U-Net [19], performs the segmentation task. The RD-Net input is the range-Doppler-channel data $RxDxCh$ while the output is a list of detections in range-Doppler with their associated class (one-hot vector). Note that for simple detection, the number of classes is two: object and not-object. In the inference stage, a global feature vector from the bottom layer of our RD-Net (512 channels) is extracted. Using spatial max-pooling on the bottom layer of the RD-Net, a global-feature vector of the size 512, is extracted.

The detections class and their locations in range-Doppler map, and a global feature vector are then passed to the Ang-Net (Fig. 6), which obtains azimuth and elevation of each detection in the range-Doppler map. For each detection, the Ang-Net pools a $3x3xCh$ crop form the input radar frame centered at the location provided by the RD-Net. The crop is then filtered with a convolution layer $3x3x256$, providing the $1x1x256$ output. This output vector is concatenated with the $1x1x512$ global feature vector and with the corresponding one-hot class vector. The concatenated vector is then passed through 3 fully-connected ($fc$) layers. The output from the fc layers is split into 2 separate classification (clc) heads: azimuth-head ($fc_4$) and elevation head ($fc_5$).

### C. Loss Functions

Similarly, to [19], the detection problem addressed in this work operates with a class-imbalanced data in which each radar data frame contains significantly larger amount of non-detection bins than any other classes. Therefore, this work uses the class-balanced cross-entropy loss in the RD-Net. The two classification heads of the Ang-net use the regular cross-entropy loss. The loss function used here is defined as:

$$L_{Tot} = L_{RD-clc} + \lambda_1 L_{Azi-clc} + \lambda_2 L_{Ele-clc}, \quad (5)$$

where $L_{RD-clc}$ is the RD-Net loss, $L_{Azi-clc}$ is the azimuth clc-head loss, $L_{Ele-clc}$ is the elevation clc-head loss and $\lambda_1, \lambda_2$ are the loss weights.

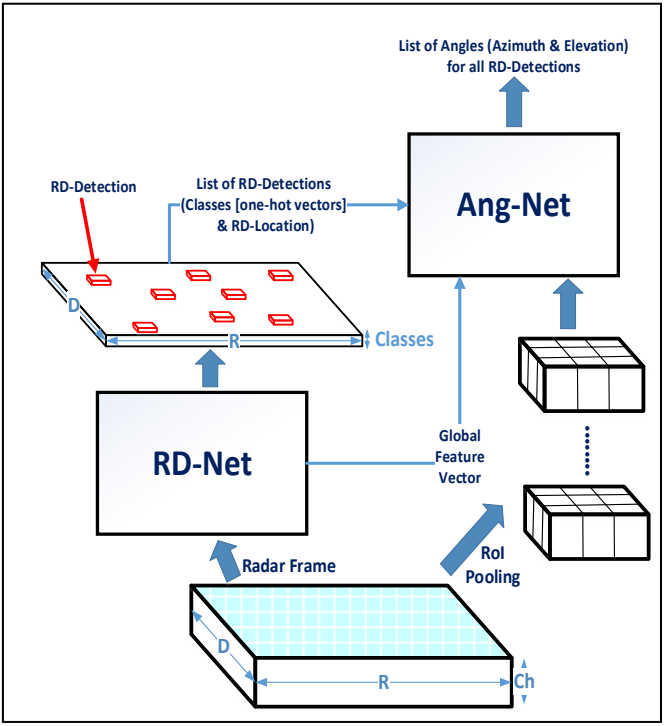

Fig. 5. DRD Network Flow. Radar frame is first passed to the RD-Net for RD-domain detection (range & doppler) and global feature extraction. The detections (location & class) are then passed to the Ang-Net, which pools for each detection a 3x3 center crop from the radar frame. It uses it with the global feature vector and class (extracted by the RD-Net) to find the angle (azimuth & elevation) of each detection.

## IV. PERFORMANCE EVALUATION

### A. Dataset Experiments

The considered model is first evaluated using 77GHz automotive radar measurements collected during the calibration procedure in the anechoic chamber. In total, 263 different radar sensors are used, and multiple radar frames with various target positions have been recorded. The data set of raw radar measurements has been constructed by a random selection of 100 data frames for each of the tested radars. In total, the considered dataset contains 26300 data frames. The dataset is split as [0.6, 0.1, 0.3] to include 15800 measurements from 158 radars for training, 2700 measurements from other 27 radars for validation, and 7800 measurements from other 78 radars for testing.

Next, following the proposed data augmentation process in section III, the raw radar measurements of each mini-batch are randomly augmented to enrich the data set with a variety of target's ranges and Dopplers. In total, after augmentation is done, we remain with 26300 radar frames but with higher diversity than the original ones. Our dataset of the radar frames contains only a single point target (corner reflector). Therefore, the 2-class network with object and non-object outputs is considered in this work.

We train all our network models with the Adam optimizer [20], with initial learning rate 0.001, betas 0.9, 0.999, weight decay 0.0005 and dropout of $p = 0.2$ in $fc1$. For the RD-Net, we use a batch size of 15 frames. For the implementation, we use PyTorch.

The training process is initiated with 5 epochs of training only the RD-Net. Following that, RD-Net and Ang-Net are trained simultaneously. During the training, the RD-Detection ground truth (GT) is used to know which of the 3x3 center crops should be passed to the Ang-Net training. For RD detections located at the edges of the RD-Map, one-sided 3x3 crop (uncentered) is used. A learning rate decay of $\gamma = 0.1$ is used in the learning steps of [5000, 60000, 100000] and the training process is executed for 250 epochs. Various $\lambda_1, \lambda_2$ in (5) are evaluated, and $\lambda_1, \lambda_2 = 1$ are selected. A threshold of 0.8 is used at the RD-Net output to filter valid detections.

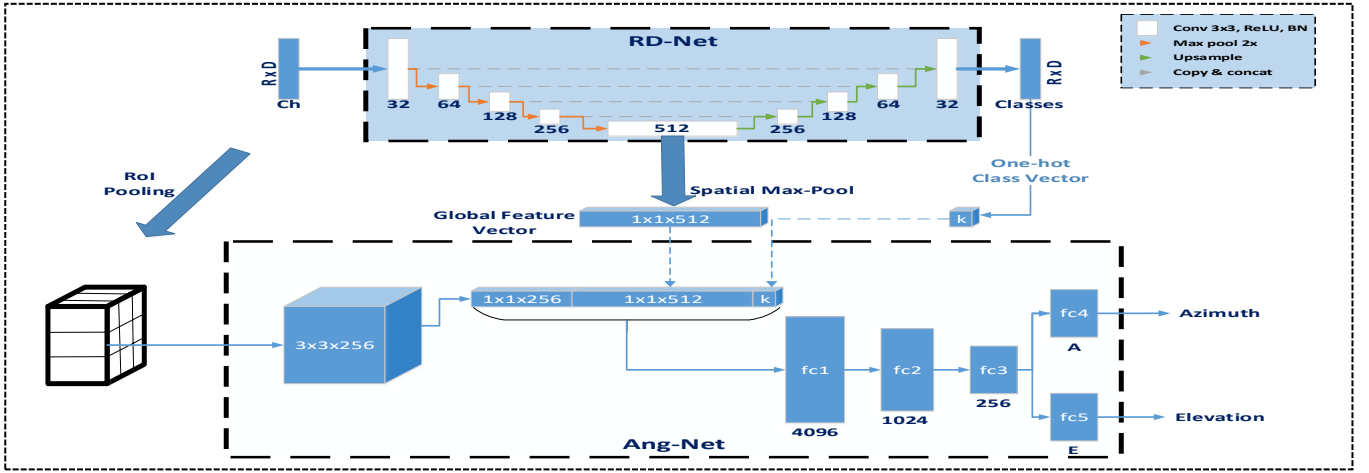

Fig. 6. DRD - Network Arcitecture. In the RD-Net a U-Net shaped network is used to "detect" all targets in the RD domain. In the Ang-Net for each detection the network takes a 3x3xCh crop and filter it with a 3x3x256 Conv resulting with a 1x1x256 vector. The vector is concatenated with the 1x1x512 global feature vector extracted from the RD-Net and also with the class one-hot vector k. The concatenated vector is then passed through 3 $fc$ layers ($fc_{1-3}$) and the output is split to 2 separate classification heads, one for azimuth detection and the second for elevation detection.

The following range and Doppler error (6) and accuracy (7) metrics are used to evaluate the methods performance:

$$R_{err} = |R_{det} - R_{gt}|, \quad (6)$$

$$D_{err} = |D_{det} - D_{gt}|,$$

$$RD_{accuracy} = \quad (7)$$
$$100 * \frac{1}{N_{det}} \sum_i (R_{err}(i) \leq 1) \text{ and } (D_{err}(i) \leq 1),$$

where $R_{det}$ and $D_{det}$ are the range and Doppler of the obtained detection, and $R_{gt}$ and $D_{gt}$ are the range and Doppler of the ground truth, $i = 1, \ldots, N_{det}$ is the index of the detection, and $N_{det}$ is the total number of detections. The accuracy threshold is set to $\pm 1$ around the detection, which means that the detection is considered as valid if it is within a 3x3 crop centered around the ground truth detection position in the range-Doppler map. Similarly, the errors and accuracy of the azimuth and elevation estimation of the detection are defined as follows:

$$Az_{err} = |Az_{det} - Az_{gt}|, \quad (8)$$

$$El_{err} = |El_{det} - El_{gt}|,$$

$$Az_{accuracy} = 100 * \frac{1}{N_{det}} \sum_i (Az_{err}(i) \leq 2), \quad (9)$$

$$El_{accuracy} = 100 * \frac{1}{N_{det}} \sum_i (El_{err}(i) \leq 2),$$

The performance of the proposed DRD approach is compared with the conventional 2D-CA-CFAR [9] and Bartlett BF [8].

For the detection in the range-Doppler map, the frame noise level ($NF$) is first estimated from the cells without the target, and the detection threshold is set to $10dB$. Thus, all cells in the range-Doppler map with energy exceeding $NF + 10dB$ are passed to the 2D-CA-CFAR. The 2D-CA-CFAR is evaluated with two sets of parameters. The first ("classic 1") uses $window\_size = 5$ and $guard\_win\_size = 1$ while the second ("classic 2") uses $window\_size = 10$ and $guard\_win\_size = 3$. For both sets we demand that the $CUT > (Avg\_Window\_Eng + 16dB)$.

For the azimuth and elevation angles, an estimate of the calibration matrix averaged over 158 train radar calibration matrices is used by the Bartlett beamformer in the test-set evaluation.

Table 1 compares the performance of the proposed DRD approach with the conventional 2D-CA-CFAR using the two considered parameters sets. Notice that the proposed DRD outperforms the conventional method. This result is even more significant, since the proposed DRD method:

- Operates in real time, processing a radar frame in ~20ms (50FPS)
- Does not require any time-consuming calibration process for each radar separately.
- Can perform in addition a classification task without any additional computational overhead.

TABLE I. PERFORMANCE COMPARISON OF DRD TO CLASSICAL RADAR APPROACH IN VARIOUS METRICS

| Metric | Method | | |
|---|---|---|---|
| | *DRD* | *Classic 1* | *Classic 2* |
| $RD_{accuracy}$ | **99.65385** | 96.61538 | 98.69231 |
| $Az_{accuracy}$ | **97.46154** | 87.71795 | 89.65385 |
| $El_{accuracy}$ | **93.01282** | 73.24359 | 73.05128 |

### B. Robustness to Noise

The data we used to train our model is originated from a calibration process, hence it has high SNR. To validate that our DRD is robust to the presence of noise we preform the following experiment. We continue to train our trained DRD network from experiment A., for extra 200 epochs using the same train parameters except the initial learning rate, which changes to 0.0001. In addition, we add uniformly random noise of 0-40[dB] to each radar frame. The accuracy results of the trained net in various SNR are shown in Fig. 7. It can be clearly observed that our DRD method is robust to the presence of noise and continues to outperform classical radar processing in various SNR conditions.

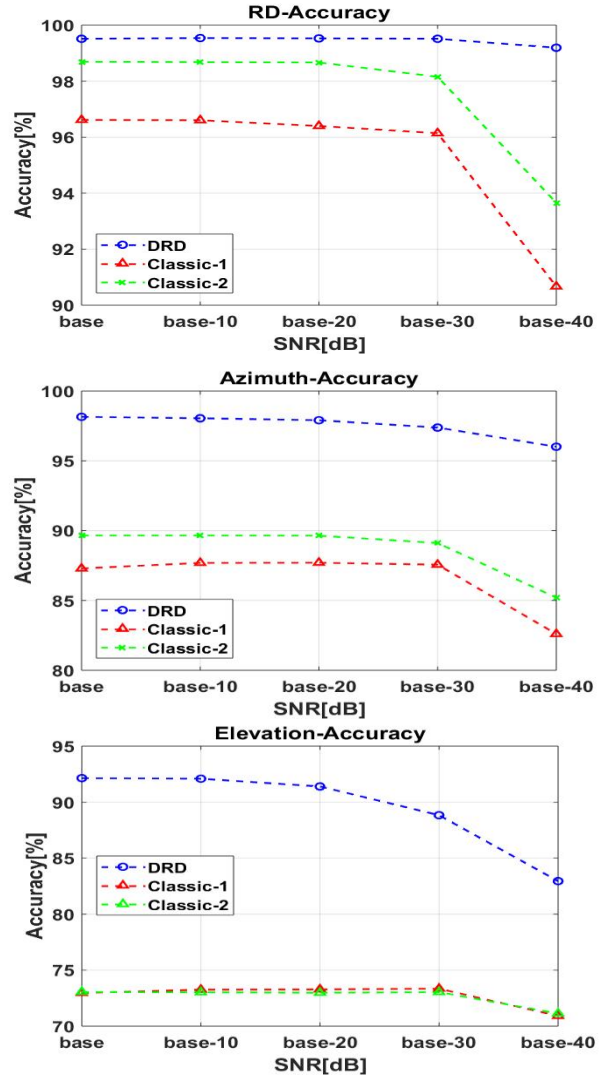

Fig. 7. Accuracy vs SNR: Range Doppler accuracy (top), Azimut Accuracy (middle), Elevation accuracy (bottom).

TABLE II. PERFORMANCE COMPARISON OF DRD VS. ANG-NET

| Metric | Method | |
|---|---|---|
| | DRD | Separate Ang-Net |
| $Az_{accuracy}$ | 97.46154 | 95.18519 |
| $El_{accuracy}$ | 93.01282 | 87.51852 |

*C. Ablation Experiment*

We recap our architectural decision to pass the global feature vector and the classes one-hot vectors from the RD-Net to the Ang-Net. To investigate the need of passing those arguments we perform the following experiment: We separately train our Ang-Net using 3x3 GT center crops, and we skip the concatenation part and pass just the feature vector of size 256 (the 3x3 Conv output) to the 3 $fc$ layers. The classification layers and the loss functions remain the same.

The network is trained using the Adam optimizer with initial learning rate of 0.01, betas of 0.9, 0.999, weight decay of 0.0005, dropout of $p = 0.2$ in $fc1$, a batch of size 40, and 3x3 crops.

Table 2 shows the azimuth-elevation accuracy of the proposed DRD approach and the separately trained Ang-Net. Notice the significant improvement in the azimuth-elevation estimation accuracy of the DRD compared to the separately trained Ang-Net. This performance improvement may be attributed to the additional information in the global features vector, which has wider field-of-view and deeper features compared to those of the 3x3 Ang-Net, and therefore we are able to finetune the network weights to better match the input radar frame.

Finally, we notice that separate Ang-Net performance are also superior to those of the classic approach, meaning we can employ our Ang-Net standalone as an alternative to classical beam-forming algorithms.

V. CONCLUSIONS

This work proposes a novel DRD approach for radar-based target detection in the 4D space of range-Doppler-azimuth-elevation. Using collected radar measurements, we show that DRD can operate in real time (~20ms for inference), outperform the classical radar methods and eliminates the need for time-consuming radar calibration. The DRD framework is suitable for multi-class classification with no additional computation overhead. The DRD is trained using the augmented radar data collected during the calibration process via a novel proposed augmentation approach.

Though this work did not include real world radar data with all the impairments that comes along, we believe that it demonstrates the potential of applying DL to radar data in general and specifically to radar complex data. The DRD method is not restricted to automotive radar and can be used for all kinds of radars.

In our future work, we plan to extend DRD detection and classification to multiple automotive radar target objects, which belong to different classes in real world scenarios. This should be done using both simulated data and real-world data automatically annotated in methods similar to [21].

ACKNOWLEDGMENT

We would like to thank Gonen Barkan for his contribution in accruing the dataset, and Erez Ben-Yaacov, Oren Longman and Hillel Sreter for providing helpful feedback throughout the work.